\documentclass[final]{cvpr}

\usepackage{times}
\usepackage{epsfig}
\usepackage{graphicx}
\usepackage{amsmath}
\usepackage{amssymb}
\usepackage{tabularx}
\usepackage{enumitem}

\usepackage[pagebackref=true,breaklinks=true,colorlinks,bookmarks=false]{hyperref}

\def\confYear{CLVISION 2021}

\begin{document}

\title{IB-DRR \-- Incremental Learning with Information-Back \\ 
Discrete Representation Replay} 

\author{
{Jian Jiang \hspace{1.1cm} Edoardo Cetin \hspace{1.1cm} Oya Celiktutan}\\
Centre for Robotics Research, Department of Engineering, King’s College London, UK \\ 
{\tt\small \{jian.jiang, edoardo.cetin, oya.celiktutan\}@kcl.ac.uk}
}

\maketitle

\begin{abstract}
Incremental learning aims to enable machine learning models to continuously acquire new knowledge given new classes, while maintaining the knowledge already learned for old classes. Saving a subset of training samples of previously seen classes in the memory and replaying them during new training phases is proven to be an efficient and effective way to fulfil this aim. 
It is evident that the larger number of exemplars the model inherits the better performance it can achieve. However, finding a trade-off between the model performance and the number of samples to save for each class is still an open problem for replay-based incremental learning and is increasingly desirable for real-life applications. In this paper, we approach this open problem by tapping into a two-step compression approach.  
The first step is a lossy compression, we propose to encode input images and save their discrete latent representations in the form of `codes' that are learned using a hierarchical Vector Quantised Variational Autoencoder (VQ-VAE). In the second step, we further compress `codes' losslessly by learning a hierarchical latent variable model with bits-back asymmetric numeral systems (BB-ANS). To compensate for the information lost in the first step compression, we introduce an Information Back (IB) mechanism that utilizes raw exemplars for a contrastive learning loss to regularise the training of a classifier. By maintaining all seen exemplars' representations in the format of `codes', Discrete Representation Replay (DRR) outperforms the state-of-art method on CIFAR-100 by a margin of 4\% average accuracy with a much less memory cost required for saving samples. Incorporated with IB and saving a small set of old raw exemplars as well, the average accuracy of DRR can be further improved by 2\%. 

\end{abstract}

\section{Introduction}
Deep neural networks leveraging large-scale annotated datasets have been shown to be powerful in many real-world tasks such as image classification. One downside is that these data-driven methods work under the assumption that all training samples (exemplars) are simultaneously available during the training phase~\cite{LR_DNN_guo2016,LR_DNN_lecun2015,EWC}. However, due to the growing need for systems that can adapt to dynamic environments and can continually learn new tasks, current deep neural networks are not adequate as they suffer from catastrophic forgetting \-- when a model is continuously updated using novel incoming data, the updates can override knowledge acquired from previous classes. Incremental learning, also known as continual learning, never-ending learning or life-long learning, aims to design systems that can keep learning new knowledge while maintaining the performance for the previously learned tasks.

\begin{figure}
\label{fig:IB}
\centering
\includegraphics[width=8.5cm]{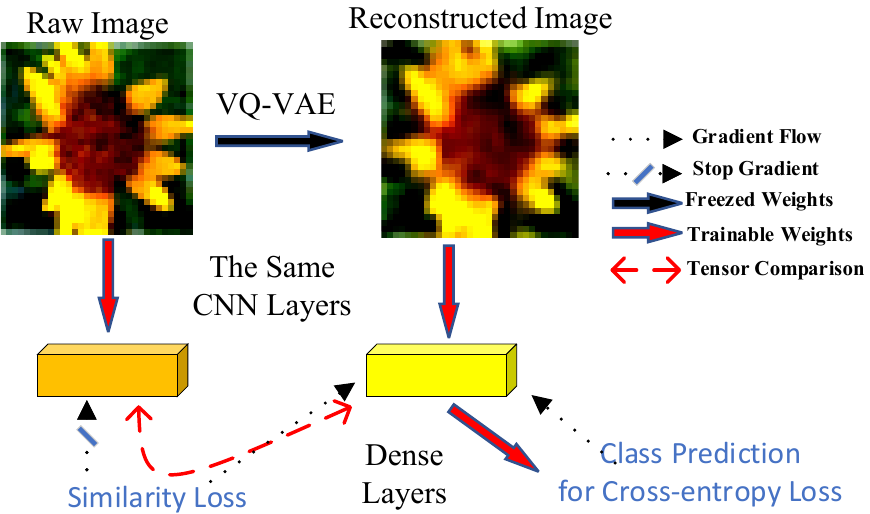}
\caption 
{ Information Back (IB) mechanism inside our classifier. IB maximizes the similarity between the latent representations of the raw image and its reconstructed view. With the help of IB, raw images are indirectly used for regularizing the optimisation of classification loss, i.e., the cross-entropy loss in our case.
}
\end{figure}
An effective and commonly used strategy in the current incremental learning methods is \textit{replay} or~\textit{rehearsal}, which is mixing the new data with the old data when learning new information to maintain old knowledge. This can be achieved by saving a small subset of old data~\cite{CL_Mnemonics,CL_ER,CL_GEM,CL_SER,CL_iCaRL,CL_LUCIR,CL_RP_LATENT}. However, the success of this strategy highly depends on how many and which exemplars to save~\cite{CL_Mnemonics,CL_iCaRL,CL_SER}. Alternatively, a line of research has focused on learning a generator to produce pseudo old samples~\cite{CL_GFR,CL_DGR,CL_VAE2020}. However, the generator needs to be updated using fake old data that may lead to a vicious cycle if some pseudo data of bad quality is produced. Recent work~\cite{CL_GFR,CL_DGR,CL_VAE2020} has shown that image generator is unstable and unreliable in incremental learning for complex dataset such as CIFAR-100~\cite{Cifar100} and only works on small and relatively simple datasets such as MNIST \cite{DS_MNIST}. 

Although carefully selected exemplars or generated representative exemplars can help improve the performance up to some degree~\cite{CL_iCaRL,CL_SER}, the larger number of exemplars a model inherits the better performance it can achieve. However, saving all available data is memory-expensive. The trade-off between model performance and the number of exemplars per class is hard to be optimised due to different applications requirement and varies among different distinctive tasks.

In this paper, we approach this open problem by introducing \textit{Discrete Representation Replay} (DRR). The DRR module allows saving the maximum number of exemplars possible at a small memory cost with two-step compression. As shown in Figure~\ref{fig:two_step_compression}, DRR benefits from a Vector-Quantised Variational Autoencoder (VQ-VAE)~\cite{VQ_VAE,VQ_VAE2} as well as a Bits-Back Asymmetric Numeral Systems (BB-ANS) that paired with hierarchical latent variables. The VQ-VAE reduces the memory required for $50,000$ uncompressed training exemplars on the CIFAR-100~\cite{Cifar100} dataset from $146.48$ MB to $7.26$ MB and BB-ANS further reduces it to $4.24$ MB, resulting in an overall memory gain of 97\%. 
To learn new classes, DRR combines latent representations (in the format of \textit{codes}) of new training samples with those of old samples and reconstruct them back to the RGB image domain at run time. Our experiments on CIFAR-100~\cite{Cifar100} showed that \textit{Discrete Representation Replay} (DRR) outperformed the state-of-the-art replay-based method~\cite{CL_Mnemonics} in the multi-class incremental learning setting by a margin of 4\% while reducing the memory size required for saving exemplars.
To alleviate the loss of information caused by VQ-VAE, we design a negative-pairs-free contrastive learning component called \textit{Information Back} (IB) inside our classifier that saves and uses raw images to regularize the training of the classifier, see Fig.~\ref{fig:IB}. We assume the reconstruction processes of different images share similar transformation patterns as the same VQ-VAE is applied. Note that IB requires to save an additional subset of raw previously seen samples but they are not directly used for minimizing the classification loss. They serve as regularizers to force the classifier to get back some general information lost during image reconstruction. Incorporated with IB component, called IB-DRR, the testing accuracy of DRR increases by 2\% on CIFAR-100. Additionally, our empirical results show that even~\textit{without maintaining previously seen raw data} IB-DRR can still boost performance in some settings, especially, the settings where the number of new classes to learn is large. In summary, our main contributions are as follows: 
\begin{itemize}[leftmargin=*]
\item We formalize incremental learning as a representation compression problem and introduce \textit{Discrete Representation Replay} (DRR) strategy for this task. More explicitly, we propose a two-step compression which allows us to save training exemplars at a competitively small memory cost. This is enabled by a VQ-VAE that is used to save and reconstruct images during new training phases and hierarchical latent variable models with BB-ANS that further reduces the memory cost. 
\item We introduce an Information Back (IB) mechanism that utilizes a subset of raw images for old classes and all raw images from new classes via a contrastive learning loss to alleviate the information loss and regularise the training. 
\item Finally, we show that on the CIFAR-100 dataset~\cite{Cifar100} our DRR surpasses the state-of-the-art method~\cite{CL_Mnemonics} by a margin of 4\% without saving any raw examples. And by saving a small number of raw samples, IB-DRR yields the best performance with a margin of 6\% as compared to the state-of-the-art method~\cite{CL_Mnemonics}. Our preliminary results on the ImageNet~\cite{DS_Downsampled_ImageNet} also indicate the viability of this approach across different datasets. 
\end{itemize}

\section{Related Work}
\label{sec:Related_work}
There is a significant body of work on incremental learning, which can be divided into three broad categories, namely, regularization-based (e.g.,~\cite{EWC,SI2017}), parameter-isolation-based (e.g.,~\cite{AE_CL,GWR}) and replay-based (e.g.,~\cite{CL_ER,CL_SER,GEM,AGEM,CL_Mnemonics,CL_GFR}). Our proposed approach falls into the category of replay-based methods. Replay, also known as rehearsal, is the most competitive and popular strategy in incremental learning and achieves state-of-the-art performance on many benchmark datasets including CIFAR-100~\cite{Cifar100}, MNIST~\cite{DS_MNIST} and ImageNet~\cite{DS_ImageNet}. We summarise these replay-based methods in three categories, namely, image-level replay, feature-level replay, and compression for replay. 
\\{\bf \textit{Image-level Replay}.} Storing and replaying exemplars in the memory to solve catastrophic forgetting was first introduced in 1990s~\cite{CL_FIRST}. Recently, Isele~\etal~\cite{CL_SER} reintroduced replay strategy in incremental learning and proposed a ranking function to select and store exemplars for each class in a long-term memory called \textit{episodic memory}. To control the memory required for replay, Rebuffi~\etal~\cite{CL_iCaRL} proposed a method name iCaRL that used a fixed size memory for all classes. To maintain the most important exemplars, they used a buffer-reconstruction procedure called \textit{Herding} to select a number of exemplars from a new class to save and discard the least preferred exemplars from previous classes to prevent memory from growing. To address the imbalance between a small number of exemplars for old classes and substantially more samples of new classes, Hou~\etal~\cite{CL_LUCIR} proposed a method named (LUCIR) that treated both old and new classes uniformly by introducing cosine normalization on latent representations of samples of new classes and old classes. They also selected samples to save by leveraging an online mining method. Most recently, Liu~\etal~\cite{CL_Mnemonics} developed a method called Mnem, where exemplars to save were considered as learnable parameters. The parameterized exemplars of old classes were optimised in an end-to-end manner during the training, and by utilizing strategies used in \cite{CL_LUCIR} their model achieved state-of-the-art performance as compared to existing replay-based methods on both CIFAR-100 and ImageNet. On the other hand, using exemplars directly for training classifiers may be prone to over-fitting~\cite{GEM,AGEM}. Therefore, Gradient Episodic Memory (GEM)~\cite{GEM} was proposed to indirectly utilize saved exemplars. Their model computed gradients given new data as well as a set of task-specific gradients using episodic memory, and it projected the gradients in the feasible region defined by task gradients. Like GEM, our \textit{Information-Back Discrete Representation Replay} (IB-DRR) also used saved raw exemplars indirectly for classification. To avoid saving original inputs, a number of works~\cite{CL_DGR,CL_VAE2020,CL_GAN} utilised a generative model such as Generative Adversarial Network (GAN)~\cite{GAN} or Variational Autoencoder (VAE)~\cite{VAE} to produce fake data that is similar to data from old classes. In this way, they aimed to transfer incremental learning problem into image generating problem. However, the generative model itself needs to maintain the ability to generate fake data for old tasks, and this approach still has to deal with catastrophic forgetting that occurs in the generative model rather than the classifier. This has been demonstrated by a recent work~\cite{GAN_CL} pointing out that a GAN itself has a catastrophic forgetting problem and the training of GAN can be regarded as an incremental learning problem. In addition, fake data may not reflect the real data distribution, which can potentially lead to a vicious cycle. Therefore, training a generative model in incremental learning settings is unstable and unreliable and recent generative-replay-based methods~\cite{CL_DGR,CL_VAE2020,CL_GAN} work on comparatively simple datasets such as MNIST~\cite{DS_MNIST} and SVHN~\cite{DS_SVHN} datasets.
\\{\bf \textit{Feature-level Replay}.} Another line of work has focused on saving the latent representations instead of saving original image data or generating fake data. Indeed, in some incremental learning scenarios~\cite{CL_GFR,CL_Review_Rahaf,CL_DGR,LWF}, training data from a certain task is no longer available after a model is trained for the task. Because saving data firstly leads to privacy and security issues and secondly results in a significant increase in the memory as the model gradually learns new classes. Pellegrini~\etal~\cite{CL_RP_LATENT} pretrained a classifier on ImageNet~\cite{DS_ImageNet}. Then its low-level layers were kept and frozen to serve as a feature extractor for other datasets like Core50~\cite{DS_Core50}. They extracted and saved a small subset of latent representations (namely, the output of the feature extractor) of inputs and replayed them during new training phases. They trained and updated a classifier using both the features of new data and replayed features. Recently, Xialei~\etal~\cite{CL_GFR} combined a generative model with a feature-level replay strategy and used a GAN to produce fake latent representations and achieved competitive performance on the CIFAR-100. Their model called generative feature replay (GFR) consisted of a modified classifier that took latent representation as inputs, a feature extractor and a generator. The modified classifier was jointly trained using cross-entropy loss with the feature extractor which encodes original images to corresponding latent representations. After trained on new data, they were frozen and the generator was updated using latent representations produced by them given corresponding data. Besides, they applied the knowledge distillation technique to the feature extractor to avoid forgetting. Feature-level replay methods generally require less memory cost compared to image-level counterparts when used to save the same amount of samples. Our approach can also be regarded as a feature-level replay method. Because we save features rather than images and the saved features are replayed and reconstructed to the image domain by the decoder in our model for training a classifier.
\\{\bf \textit{Compression for Replay}.} There are three most related continual learning works~\cite{CL_EEC,CL_online_VQVAE,CL_discreteVAE} similar to our DRR. Ayub~\etal~\cite{CL_EEC} used a pretrained autoencoder structure that saved compressed latent representations of previous samples and reconstructed them back to image-domain for replaying. 
Caccia~\etal~\cite{CL_online_VQVAE} used stacked VQ-VAEs and updated the codebooks to control drifting representations during the incremental learning whereas our DRR used a hierarchical VQ-VAE and our codebooks were frozen after the VQ-VAE was pretrained. Caccia~\etal reported that saving more samples might degrade the image quality due to the drift that occurred in their VQ-VAE. Thus, in their approach storing more samples did not necessarily improve classification performance. Our results showed that our VQ-VAE with frozen codebooks did not suffer from data shift problem much. 
Matthew~\etal~\cite{CL_discreteVAE} used a discrete VAE model with a `codes' buffer. However, unlike our DRR whose latent space was able to represent infinite latent representations of samples, their latent space had a restricted size that was proportional to the VAE capacity (VAE capacity was a major hyperparameter in their work). As shown in their work, the smaller capacity their VAE had, the higher compression it could achieve but the fewer maximum samples they could save. Besides, to make their discrete VAE fit new data, the VAE was updated using reconstructed images via codes in the buffer as well as real images from new tasks. It potentially transferred the problem of catastrophic forgetting from classifiers into generators like other generative models we discussed before. In contrast, our trained DRR has a stable high compression rate with generally good reconstructed image quality and can be generalized to other datasets (see Appendix for reconstructed images via DRR). 

\section{Background}

\subsection{Vector Quantised Variational AutoEncoder}
\label{sec:VQ_VAE}
Vector Quantised Variational AutoEncoder (VQ-VAE)~\cite{VQ_VAE} was originally designed for reducing the computation workload in PixelCNN~\cite{PixelCNN} for the task of image generation. PixelCNN-based methods~\cite{PixelCNN,PixelSnail} can generate high-quality fake images, however, due to the nature of auto-regressive models, they need to be trained over each pixel leading to a considerably high computation cost. The VQ-VAE encodes an RGB image into a `codes' matrix with a smaller size. Instead of original images, the codes are then used as an input to auto-regression models, resulting in a significant computation reduction. The VQ-VAE relies on vector quantisation that encodes inputs into discrete representations rather than continuous ones and as a compressor it has been shown to be effective in image generation and image compression~\cite{VQ_VAE,VQ_VAE2}. Similarly, in this paper, we benefit from the VQ-VAE not only as a good feature extractor but also as a powerful compressor. Considering that a PixelCNN is also a generative model that has to deal with incremental learning as we discussed in Section~\ref{sec:Related_work}, instead of generating fake images using an auto-regressive generator, our model reconstructs the images via saved codes using the VQ-VAE only.
The VQ-VAE defines a discrete embedding space called `codebook' $\mathbf{c} \in \mathbb{R}^{K \times d}$, where $K$ is the size of codebook and $d$ is the dimension of each embedding vector $c_k \in\mathbb{R}^{d}, k\in1,2,3,...K$. The encoder $E(x)$ learns a non-linear mapping that encodes the input $\mathbf{x} \in\mathbb{R}^{M}$ to the latent representation $\mathbf{z_e} \in\mathbb{R}^{d}$, where $d \ll M$. Then vector quantisation $VQ(.)$ is applied over $\mathbf{z_e}$ and, given the codebook (the prototype set), $\mathbf{z_e}$ is replaced with a set of embedding vectors in the codebook using nearest neighbor search as formalised by
\begin{equation}
\label{eq:VQ_searching}
\mathbf{z}_{\mathbf{q}}=VQ\left(\mathbf{z}_{\mathbf{e}}\right)=\mathbf{c}_{\mathbf{k}}, \text { where } \mathbf{k}=\arg \min _{\boldsymbol{k}}\left\|\mathbf{z}_{\mathbf{e}}-c_{k}\right\|_{2}
\end{equation}

The quantised output $\mathbf{z}_{\mathbf{q}}$ is then fed into the decoder to obtain the reconstructed data. The objective of VQ-VAE can be defined as 
\begin{align}
\mathcal{L}(\mathbf{x}, D(\mathbf{c_k})) = &\|\mathbf{x}-D(\mathbf{c_k})\|_{2}^{2} \label{eq:reconstruction_error} \\ 
&+\|s g[E(\mathbf{x})]-\mathbf{c_k}\|_{2}^{2} \label{eq:codebook_loss} \\
&+\beta\|s g[\mathbf{c_k}]-E(\mathbf{x})\|_{2}^{2} \label{eq:commitment} 
\end{align}
where the term in Eq.~\ref{eq:reconstruction_error} is the \textit{reconstruction error} of a typical autoencoder (e.g., mean squared error). And $D(.)$ refers to the decoder of the VQ-VAE. The term in Eq.~\ref{eq:codebook_loss} is called \textit{codebook loss}, which brings the selected embedding vectors $\mathbf{c_k}$ close to $\mathbf{z_e}$. Besides, the gradients produced by Eq.~\ref{eq:codebook_loss} are applied to the codebook only. The last term in Eq.~\ref{eq:commitment}, called \textit{commitment loss}, enables encoder to produce similar values to the chosen embedding vectors and it is only applied to the encoder weights.

A hyperparatmer $\beta$ is used to balance the learning rates of different terms and $sg[.]$, namely stop-gradient, is an operation that prevents gradients to propagate to its argument. For example, in Eq.~\ref{eq:codebook_loss} $sg[E(x)]$ blocks the gradients calculated by this loss term from flowing into the weights of the encoder $E(.)$. The visualized vector quantisation operation can be found in Fig.~\ref{fig:two_step_compression}. 

\subsection{Bits-Back Asymmetric Numeral Systems}
 
Asymmetric Numeral Systems (ANS) is used to compress sequences of discretely distributed symbols (a `symbol' is a one dimension data point) into a sequence of bits and can recover those bits back to symbols. 

Bits-back with ANS (BB-ANS)~\cite{BB-ANS,BB-ANS_HILLOC,Bitswap} tries to approximate true data distribution $p_{data}(\mathbf{x})$ by $p_{\boldsymbol{\theta}}(\mathbf{x})$ and encode datapoint into a number of bits equal to its negative log probability assigned by a latent variable model. 

\begin{figure*}[htb]
\includegraphics[width=1\textwidth]{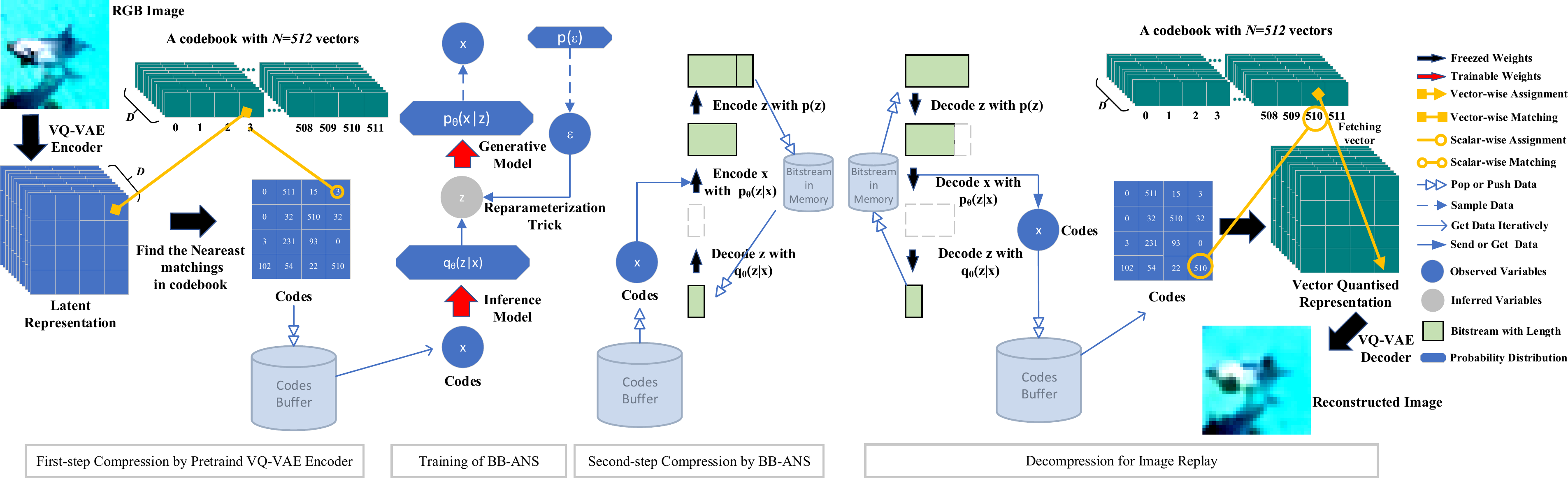} 
\caption{DRR uses a two-step compression. In the first step, a pretrained VQ-VAE encodes RGB images into codes that are saved in a buffer. During each new phase, BB-ANS is fine-tuned with codes from new images and those previous saved. Updated BB-ANS is used to perform the second step compression, i.e., compress codes and save them into bit-streams in the memory. Discrete latent representations are replayed by first decompressing bit-streams to codes and then decompressing the codes back to the image domain. This data flow only shows the core idea of DRR. In practice, a hierarchical VQ-VAE and a BB-ANS with hierarchical latent variables are used.}
\label{fig:two_step_compression}
\end{figure*}

BB-ANS models use latent variable models with a marginal distribution $p_{\boldsymbol{\theta}}(\mathbf{x})$ defined by $p_{\boldsymbol{\theta}}(\mathbf{x})=\int p_{\boldsymbol{\theta}}(\mathbf{x}, \mathbf{z}) \mathrm{d} \mathbf{z}=\int p_{\boldsymbol{\theta}}(\mathbf{x} | \mathbf{z}) p(\mathbf{z}) \mathbf{d} \mathbf{z}$, and they utilize an \textit{inference model} $q_{\boldsymbol{\theta}}(\mathbf{z}|\mathbf{x})$ of variational autoencoders (VAE)~\cite{VAE} to approximate the posterior $p_{\boldsymbol{\theta}}(\mathbf{z} | \mathbf{x})$. The marginal likelihood $ p_{\boldsymbol{\theta}}(\mathbf{x})$ is rewritten as:
\begin{equation}
\label{eq:marginal_likelihood}
\log p_{\boldsymbol{\theta}}(\mathbf{x})=\mathbb{E}_{q_{\boldsymbol{\theta}}(\mathbf{z} \mid \mathbf{x})} \log \frac{p_{\boldsymbol{\theta}}(\mathbf{x}, \mathbf{z})}{q_{\boldsymbol{\theta}}(\mathbf{z} \mid \mathbf{x})}+\mathbb{E}_{q_{\boldsymbol{\theta}}(\mathbf{z} \mid \mathbf{x})} \log \frac{q_{\boldsymbol{\theta}}(\mathbf{x}, \mathbf{z})}{p_{\boldsymbol{\theta}}(\mathbf{z} \mid \mathbf{x})}
\end{equation}

\noindent
where the first term in Eq.~\ref{eq:marginal_likelihood} is the Evidence Lower BOund (ELBO) that is jointly optimized using the reparameterization trick with the \textit{inference model} and \textit{generative model} of VAE. However, for latent variable models, to encode $\mathbf{x}$, it is necessary to encode $\mathbf{z}$ as well, inducing an extra message length for the prior $p(\mathbf{z})$ equals to $-\log p(\mathbf{z})$. To get those extra bits back, BB-ANS firstly initializes ANS with a bit-stream of $N$ random bits. Then it performs three steps (denoted as \textit{S1,S2,S3}) as follows: S1 Decode $\mathbf{z}$ from bit-stream using the \textit{inference model} $q_{\boldsymbol{\theta}}(\mathbf{z}|\mathbf{x})$; S2 Encode $\mathbf{x}$ using \textit{generative model} $p_{\boldsymbol{\theta}}(\mathbf{x}|\mathbf{z})$; S3 Encode $\mathbf{z}$ to bit-stream using the prior $p(\mathbf{z})$. The S2 and S3 add $-\log p_{\boldsymbol{\theta}}(\mathbf{x} | \mathbf{z})$ bits and $-\log p(\mathbf{z})$ respectively, but BB-ANS gets $-\log q_{\boldsymbol{\theta}}(\mathbf{z} | \mathbf{x})$ bits back from the S1. The net message length then becomes $\log q_{\boldsymbol{\theta}}(\mathbf{z} | \mathbf{x})-\log p_{\boldsymbol{\theta}}(\mathbf{x} | \mathbf{z})-\log p(\mathbf{z})$, which is on average equal to the negative ELBO. The cost of initial $N$ random bits is negligible for encoding long sequences. A visualized BB-ANS in shown in Fig.~\ref{fig:two_step_compression}. 
\section{Discrete Representation Replay (DRR)}
\label{sec:DRR}
In this section, we formalize the incremental learning as a representation compression problem, and propose \textit{Discrete Representation Replay} (DRR) by a two-step compression as shown in Fig.~\ref{fig:two_step_compression}. 
\\{\bf \textit{Lossy Compression by VQ-VAE}.}
In the first step of compression, we utilise a VQ-VAE to process the inputs and save their discrete latent representations in a format called `codes' that can be later used to reconstruct the original inputs. The VQ-VAE provides a lossy compression where `reconstructed' images in our approach are a blurrier version of the original ones \-- naturally caused by an undercomplete autoencoder that gradually reduces the dimensionality of its intermediate layers and results in a smaller size of latent representation compared to the original input. This lossy compression contributes to the majority of the total compression rate of DRR. In our proposed model, we use a hierarchical VQ-VAE introduced in~\cite{VQ_VAE2} (see Appendix for the network structure). Because our preliminary results have shown that the original VQ-VAE~\cite{VQ_VAE} cannot reconstruct the images adequately for our task. In our hierarchical VQ-VAE, vector quantisation is applied $2$ times after different intermediate layers. It aims to learn hierarchical discrete latent representations such that a top-level representation models global information while a bottom-level representation captures local details. Two levels are trained jointly inside the hierarchical VQ-VAE and two independent codebooks are used respectively. Additionally, we adopt a fully convolutional design for our VQ-VAE, i.e., all of the layers are either convolutional or element-wise functions. In this way, the VQ-VAE is able to process images with arbitrary sizes. Our experiments show that our VQ-VAE trained on CIFAR-100 can reconstruct images with a higher resolution well for classification like images from ImageNet.
After we pretrain the VQ-VAE, we freeze and use it as a feature extractor and save the resulting latent representations in the form of codes, which significantly decreases the memory cost.
\\{\bf \textit{Lossless Compression by BB-ANS}.}
In the second step, we utilize a BB-ANS with hierarchical latent variables called Bit-Swap introduced in~\cite{Bitswap} that is more expressive than original BB-ANS~\cite{BB-ANS}, i.e., can more closely model real data distribution. Instead of using a fixed prior for $\mathbf{z}$, Bit-Swap assumes that $\mathbf{z}$ is generated by a latent variable $\mathbf{z1}$, and $\mathbf{z1}$ is generated by another latent variable $\mathbf{z2}$ and so on. In this way, the sampling process of both the generative model and the inference model obeys a Markov chain dependency between the stochastic variables~\cite{Bitswap}. In a lossless manner, we use Bit-Swap to compress `codes' produced by the VQ-VAE. We train from scratch a Bit-Swap model at first and keep fine-tuning it for future new data distribution of codes.

In DRR, a new classifier is trained with reconstructed images both from new classes and old classes. Our preliminary results show that training from scratch the classifier every time it is required to learn new classes always performs better than fine-tuning the previously learned classifier. Though it is widely accepted that the learned model weights contain information of old data and the learned weights are an unwritten definition for the concept ‘past knowledge’ in the increment learning domain, intuitively, if given access to old distribution, training the model from scratch is more flexible, leading a better fit to the new data distribution (a mixture of old and new data). In fact, He~\etal~\cite{Rethinking_pretraining} show that in some cases training from scratch is more robust and usually converges a solution \textit{on par with or even better than} the fine-tuning counterpart.

\section{Information-Back DRR (IB-DRR)}
\label{section:IB-DRR}
The pseudo data distribution produced by either generative models or reconstruction models has a drift towards original real data distribution. In our DRR, we lose the information of real data because of the nature of under-complete autoencoders as well as vector quantisation operations. This causes a considerably large decrease in classification score as compared to using original data, e.g., around an overall 12\% decrease for our DRR on the CIFAR-100 dataset. We can regard the reconstructed images as augmented data that is transformed by some blurring transformation, though this kind of augmentation may not necessarily be a good augmentation for classification. To remedy this problem, we propose to utilize contrastive learning by treating an image and its reconstructed version as a positive pair.
Inspired by a recent negative-pairs-free contrastive learning contrastive learning method called SimSiam~\cite{SimSiam}, we propose an \textit{Information Back} (IB) mechanism for DRR, see Fig.~\ref{fig:IB}. Specifically, our IB-DRR maximizes the cosine similarity of the latent representations (produced by the classifier) of real images and their reconstructed views produced by VQ-VAE. We denote the convolution layers of a classifier before its dense layers (fully connected layers) as $f$, and the output of $f$, i.e., latent representations of raw input 
$x$ and its reconstructed view $\hat{x}$ as $r_{1} \triangleq f(x)$ and $r_{2} \triangleq f(\hat{x})$, respectively. The objective of IB is to minimize the negative cosine similarity between $r_1$ and $r_2$: 
\[
\label{eq:sim}
\mathcal{L}_{sim} = -sg \left[\frac{r_{1}}{\left\|r_{1}\right\|_{2}} \right]  \cdot \frac{r_{2}}{\left\|r_{2}\right\|_{2}}
\]
where $sg[.]$ prevents gradients to propagate to its argument.
Then IB-DRR jointly optimises the $\mathcal{L}_{sim}$ and a cross-entropy loss $\mathcal{L}_{ce}$, i.e., $\mathcal{L} = \lambda \mathcal{L}_{sim} + \mathcal{L}_{ce}$ where $\lambda$ is a scalar used to balance the two loss terms.
Note that IB-DRR requires maintaining two memory set, one is codes of all old training samples, the other is a small subset of raw training exemplars of old classes. Like GEM~\cite{GEM,AGEM}, the saved raw old exemplars are not directly used for minimizing the classification loss. They serve as regularizers to force the classifier to get back some lost information that is important for classification. More explicitly, it forces the classifier to treat reconstructed images and the real ones in a similar way. The general training process of IB-DRR is similar to DRR except that IB-DRR jointly optimises the two losses as follows: (1) the contrastive learning loss between raw exemplars and their reconstructed views; and (2) the cross-entropy loss computed using reconstructed images from new exemplars and old codes.

\section{Experiments}
Our experiments are designed for class-incremental classification and we consider the single-headed classifier scheme where there is only a single unified output layer. In other words, a unified classifier learns novel classes and classifies all the classes that it has seen so far. We compare our approach with baseline methods and the state-of-the-art methods that use the replay strategy.

\subsection{Experimental Setup} 
\label{subsec:Experimental Setup}
Following~\cite{CL_LUCIR,CL_Mnemonics,CL_GFR,CL_iCaRL}, we evaluate our proposed strategies on the CIFAR-100 dataset~\cite{Cifar100}, and we train three variants for ablation study: 1) DRR: discrete representation replay with saving latent representations (in the format of codes) of all seen exemplars; 2) IB-DRR: information back discrete representation replay with an extra requirement of saving a small set of raw original exemplars of previously seen classes; and 3) IB-DRR$^*$: a variant of IB-DRR without saving raw data, so it can only utilize raw (original) exemplars of new coming classes for IB. More explicitly, IB-DRR$^*$ is IB-DRR without saving raw samples for old classes; and DRR is obtained by ablating Information Back mechanism from IB-DRR$^*$.\\
{\bf \textit{Dataset}.} The CIFAR-100 dataset contains $100$ classes and each class has $500$ training samples and $100$ test samples with the image size of $32\times32\times3$. 
\\{\bf \textit{Architecture}.} For our hierarchical VQ-VAE, we use two codebooks (two level of codes) each with a size of $512$ (number of codes) and set the embedding dimension to $64$. We use a Bit-Swap with $8$ hierarchical latent variables with a Markov chain structure for each level of codes. In our preliminary experiments, we find the Resnet-32 used in previous replay-based methods~\cite{CL_Mnemonics,CL_US_CURL,CL_iCaRL} causes underfitting of our DRR in some cases. We use a ResNet-18~\cite{Resnet} used in~\cite{CL_GFR} as a classifier (see Appendix for further discussion).
\\{\bf \textit{Hyperparameters and Configuration}.} Following the \textit{class incremental} settings introduced in~\cite{CL_LUCIR,CL_Mnemonics,CL_GFR}, our VQ-VAE and Bit-Swap are pretrained given the half of the classes from CIFAR-100 at the initial training phase $0$ and the rest of the classes are added gradually in the future phases. That is, the training has $1$ initial phase for the classification of the first $50$ classes and $N$ incremental phases for the remaining $50$ classes. We freeze the VQ-VAE but keep fine-tuning the Bits-Swap for future phases. Note that our classifier is trained with reconstructed images but tested with raw images. Our preliminary results show that training the classifier with the mixture of raw images of new classes and reconstructed images of old classes degrades the classification results. 
We evaluate the performance of our incremental learning method by setting $N$ to $25$, $10$ and $5$, meaning that $2$, $5$, and $10$ classes are gradually added at each incremental phase, respectively.
The VQ-VAE is trained by an Adam optimizer with a constant learning rate $0.0003$ for $1400$ epochs. We train or fine-tune the Bit-Swap by an Adam optimizer with a constant learning rate $0.002$ until its loss no longer decreases.
For DRR, IB-DRR and IB-DRR$^*$, we train the classifier with the image replay strategy as explained in Section~\ref{sec:DRR}. We use a stochastic gradient descent (SGD) optimizer with $0.9$ momentum and $0.0005$ weight decay parameters for the classifier. The initial learning rate is set to $0.1$ and divided by $5$ after $60$, $120$, and $160$ epochs. A warmup scheduler \cite{warmup} is applied for the first $5$ epochs. The classifier is trained from scratch during a new training phase for $200$ epochs. If IB is used, we set the hyperparameter $\lambda = 0.005$ for the contrastive learning loss $\mathcal{L}_{sim}$. \\
{\bf \textit{Memory Budget}.} For DRR and IB-DRR$^*$ we only save codes for seen classes. For IB-DRR, we follow~\cite{CL_Mnemonics,CL_US_CURL,CL_iCaRL} and save additionally $20$ raw exemplars per class for seen classes.\\
{\bf \textit{Baselines}.} We compare our methods with the following 5 baseline methods: {\bf \textit{LWF}}~\cite{LWF}, {\bf \textit{GFR}}~\cite{CL_GFR}, {\bf \textit{iCaRL}}~\cite{CL_iCaRL}, {\bf \textit{LUCIR}}~\cite{CL_LUCIR}, {\bf \textit{Mnem}}~\cite{CL_Mnemonics} and a {\bf{Upper Bound (UB)}}. The first two baselines do not require to save exemplars while others (including ours) need to save old exemplars. iCaRL, LUCIR and Mnem save $20$ samples per class while UB saves all (50,000) samples. We follow their original implementation settings (see Appendix for further information).

\subsection{Evaluation Metrics}
For each new training phase $i$, we evaluate the performance of the model on the test set from the new classes and old classes (henceforth, phase-wise accuracy). We use two evaluation metrics. Following the previous work~\cite{CL_LUCIR,CL_GFR,CL_iCaRL}, the first metric is the \textit{average overall accuracy} that computes the mean of phase-wise accuracy over all training phases, namely from the initial phase to phase $i$. However, in the initial phase $N_0$, there is no catastrophic forgetting; 
therefore, we exclude the test accuracy of the initial phase from the \textit{average overall accuracy}. Let $A_{i}$ be the performance of the model on the held-out test set of phase $i$, the average overall accuracy is then defined as:
$
\overline{\mathcal{A}}=\frac{1}{N} \sum_{i=1}^{N} A_{i}
$
where $N$ is the total number of phases (in our experiments $N=5, 10$, and $25$). The second metric is the \textit{last phase accuracy} that reports the phase-wise accuracy of the last training phase only. Please note that, in the last phase, all classes in the dataset are encountered, thus the accuracy is given for all $100$ classes. 
\begin{table}[t]
    \centering 
    \begin{tabularx}{.45\textwidth}{l c c c } 
    \hline
    Method  & 25 phases & 10 phases & 5 phases \\ [0.5ex] 
    \hline 
    \textit{UB} & \textit{78.60} & \textit{78.92} & \textit{78.86} \\
    GFR \cite{CL_GFR} &54.01 & 60.14 & 60.18\\
    Mnem \cite{CL_Mnemonics} & 60.96 & 60.78 & 60.76\\
    DRR  & 64.37 & 64.04 & 64.00 \\
    IB-DRR$^*$  & 62.90 & 64.85 & 65.36 \\
    IB-DRR  & \bf{65.65} & \bf{66.40} & \bf{66.71} \\
    \hline 
    \end{tabularx}
    \caption{Comparison in terms of the average overall accuracy (\%) for varying $N$ incremental phases on CIFAR-100.} 
    \label{tab:SOTA_avg_acc} 
\end{table}
\begin{table}[t]
    \centering 
    \begin{tabularx}{.45\textwidth}{l c c c } 
    \hline
    Method  & 25 phases & 10 phases & 5 phases \\ [0.5ex] 
    \hline 
    \textit{UB} & \textit{77.91} & \textit{77.91} & \textit{77.91}\\
    
    GFR  \cite{CL_GFR} & 40.39 & 51.29 & 53.34\\
    
    Mnem \cite{CL_Mnemonics} & 50.78 & 51.53 & 54.32  \\
    DRR & 60.87 & 60.87 & 60.87  \\
    IB-DRR$^*$ & 60.50 & 62.29 & 62.83  \\
    IB-DRR  & \bf{62.86} & \bf{63.65}  & \bf{63.97} \\

    \hline 
    \\
    \end{tabularx}
    \caption{Comparison in terms of last phase accuracy (\%) (after models trained for 100 classes) on CIFAR-100 for varying $N$ incremental phases.}
    \label{tab:SOTA_last_acc} 
\end{table}
\subsection{Results and Analyses}
\begin{figure*}[htb]
\includegraphics[width=1\textwidth]{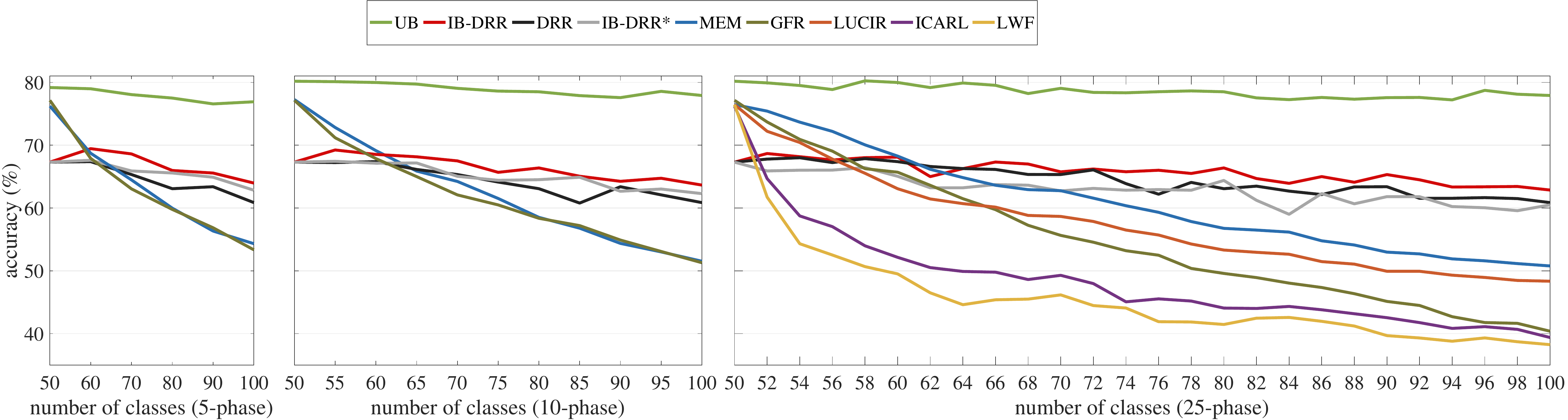} 
\caption{Phase-wise accuracy (\%) on Cifar100 based on 5-phase (10 class per phase), 10-phase (5 class per phase), and 25-phase (2 class per phase) settings, respectively.}
\label{fig:SOTA_3setings}
\end{figure*}

In Figure~\ref{fig:SOTA_3setings}, we compare our methods with the state-of-the-art replay-based methods that use image-level replay~\cite{CL_Mnemonics} and feature-level replay~\cite{CL_GFR} as well as other baseline methods~\cite{CL_iCaRL,LWF,CL_LUCIR} in terms of \textit{phase-wise accuracy}. It is obvious that methods using replay have better performance than LwF~\cite{LWF} which is one of the most competitive methods without replay, demonstrating the superiority of replay-based methods. IB-DRR outperforms others except for the upper bound (UB) in all experimental settings.

Looking at the most challenging $25$-phase setting, our models start with a lower phase-wise accuracy at the initial training phase (with $50$ classes) due to the fact that reconstructed images are of low quality as compared to original images. However, despite the fact it is trained with reconstructed, low-quality images, our models can maintain the knowledge over old classes when learning new classes with improvement in the average accuracy by a margin of $6\%$ and $4\%$ by IB-DRR and DRR respectively (see Table~\ref{tab:SOTA_avg_acc}). In addition, IB-DRR and DRR surpass other baselines by a margin of at least $9\%$ and $3\%$ according to the \textit{last phase accuracy} (see Table~\ref{tab:SOTA_last_acc}). In $5$-phase and $10$-phase settings, IB-DRR$^*$ performs better than DRR while in $25$-phase setting we see the opposite situation. Recall that, both of them only save codes for previously seen classes. Intuitively, the Information Back (IB) mechanism benefits more diverse raw samples from more new classes to regularize the classifier to get back some information that is useful for classification. From this point of view, in $5$-phase or $10$-phase IB-DRR$^*$ has a large enough number of new classes to utilize. As IB-DRR also saves old raw samples, it can perform well in all the settings we tested. Moreover, our methods behave similarly to the upper bound with a relatively `constant' decrease in accuracy.

\begin{table*}[t]
    \centering 
    \begin{tabular}{l c c } 
    \hline
    Method  & Exemplar cost (MB) & Assistant model cost (MB)  \\ [0.5ex] 
    \hline 
    UB  & 148 (50,000 exemplars)  & NA \\
    GFR~\cite{CL_GFR} & NA &  4.5 (GAN)\\
    Mnem~\cite{CL_Mnemonics} or LUCIR~\cite{CL_LUCIR}  & 6.2 (2,000 exemplars)  & NA\\ 
    
    DRR or IB-DRR$^*$  & 4.24 (50,000 exemplars) & 5.81 (VQ-VAE) + 13.4 (Bit-Swap)\\
    IB-DRR  &  6.2 + 4.24 & 5.81 (VQ-VAE) + 13.4 (Bit-Swap)\\
   
    \hline 
    \\
    \end{tabular}
    \caption{Comparison of memory required for saving exemplars and assistant models, e.g., generators and autoencoders on CIFAR-100.} 
    \label{tab:SOTA_memory} 
\end{table*}
\subsubsection{Comparison of Memory Cost}
\label{subsec:memory}
Table~\ref{tab:SOTA_memory} provides the memory required for saving extra exemplars as well as assistant model weights on the CIFAR-100. Saving discrete latent representations in codes costs less memory while saving 25 times more exemplars as compared to other replay-based methods~\cite{CL_Mnemonics,CL_LUCIR}. 

Please note that we are not able to compare the cost of assistant models. Because many incremental learning methods (especially those that use knowledge distillation techniques) save a copy of old models. For example, Mnem~\cite{CL_Mnemonics} saves the model weights $\Theta_{0}$ after the $1^{st}$ phase and uses knowledge distillation loss to encourage $\Theta_{i-1}$ and $\Theta_{i}$ to maintain the same prediction ability on old classes. GFR~\cite{CL_GFR}, LUCIR~\cite{CL_LUCIR} and Mnem~\cite{CL_Mnemonics} all need to save a copy for the latest model and train a new model with the help of the old model copy, therefore, each of them should be part of the total memory cost of assistant models. 

\subsection{Can DRR generalise to other datasets?}
\label{sec:generalization_VQVAE}
To validate that our VQ-VAE trained on CIFAR-100 has a good generalization ability when applied to a different dataset even with a higher resolution,
we presented preliminary quantitative results on Subset-ImageNet that contained random $100$ classes from ImageNet-1K. For our method, we used \emph{the pretrained VQ-VAE on CIFAR-100} as described previously in Section~\ref{subsec:Experimental Setup} to process images from ImageNet. We compared our DRR with Mnem~\cite{CL_Mnemonics} which is the SOTA method. Both approaches used images with the size of $224\times224\times 3$ as inputs and Resnet-18 as the classifier. Mnem saved $20$ raw samples per class ($330$ MB in total) while DRR saved codes of all samples.
Results of the $5$-phase setting showed DRR and Mnem achieved an average overall accuracy of {\bf{$78.05\%$}} and $72.58\%$ respectively. In addition, DRR resulted in a memory cost of $474$ MB for saving $128,856$ raw exemplars which originally cost $18,494$ MB. 
\section{Conclusion}
In this paper, we formalised incremental learning as a representation compression problem and proposed a novel approach to this problem. Our proposed approach, Discrete Representation Replay (DRR), performs a two-step compression using a Vector-Quantised Variational Autoencoder (VQ-VAE) and a Bits-back Asymmetric Numeral Systems with hierarchical latent variables (Bit-Swap). 
Our experimental results showed that DRR outperforms the state-of-the-art approaches on the CIFAR-100 dataset in terms of accuracy and memory size required for saving exemplars. In addition, we introduced an Information Back (IB) mechanism that utilized raw exemplars to regularize the training of the classifier and IB further boosted the performance of the DRR by saving a small set of raw exemplars of previously seen classes. 
Our preliminary results showed implications that our approach could be generalised to other datasets such as ImageNet. As future work, we will extend IB-DRR by introducing extra network components for contrastive learning to alleviate information loss caused by the VQ-VAE.
{\small
\section*{Acknowledgement}
\noindent
The work of Jian Jiang has been supported by the King's China Scholarship Council (K-CSC) PhD Scholarship programme. 
}
{\small
\bibliographystyle{ieee_fullname}
\bibliography{egbib}
}
\newpage
\appendix
\section{Qualitative Results of VQ-VAE in CIFAR-100 and ImageNet}
\begin{figure}
\centering
\includegraphics[width=8.5cm]{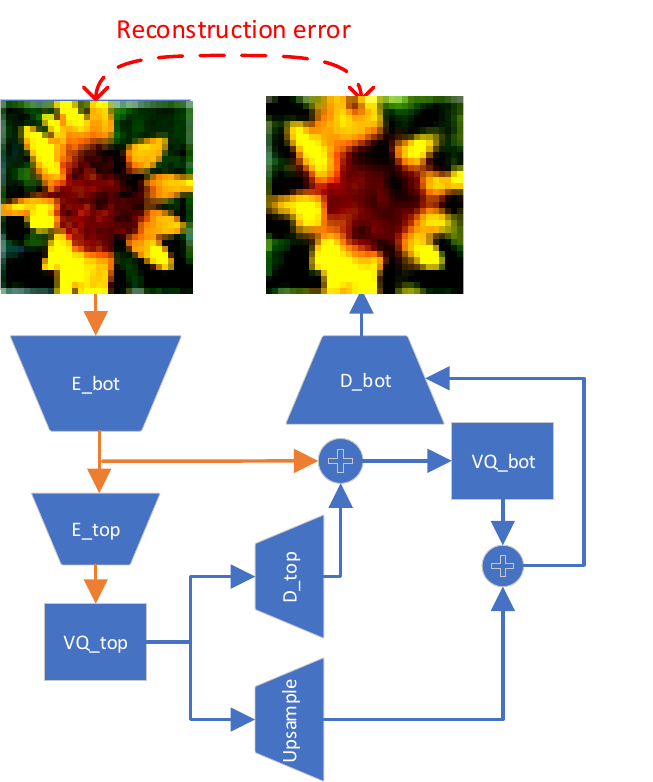}
\caption {The hierarchical VQ-VAE architecture for incremental learning. E\_, D\_, VQ\_ represent Encoder, Decoder, and Vector quantisation respectively.}
\label{fig:VQ-VAE2}
\end{figure}

In this section, we show the qualitative results of reconstructed images by the hierarchical VQ-VAE pretrained given half classes of CIFAR-100 (the one used in our main paper). Fig.~\ref{fig:VQ-VAE2} shows the model architecture of the hierarchical VQ-VAE.

\begin{figure*}[htb]

\begin{tabular}{c}
        \includegraphics[width=.9\textwidth]{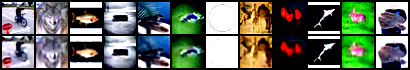} \\ 
         (a) Reconstructed images from CIFAR-100 with resolution $32\times 32$ \\
        \includegraphics[width=.9\textwidth]{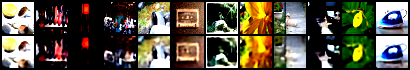}  \\ 
         (b) Reconstructed images from down-sampled ImageNet with resolution $32\times 32$ \\
      \includegraphics[width=.9\textwidth]{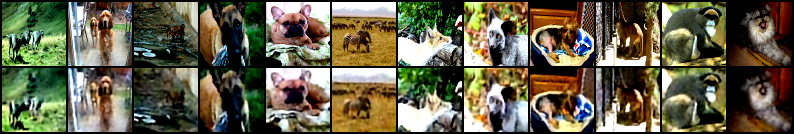}  \\
         (c) Reconstructed images from down-sampled ImageNet with resolution $64\times 64$\\
       \includegraphics[width=.9\textwidth]{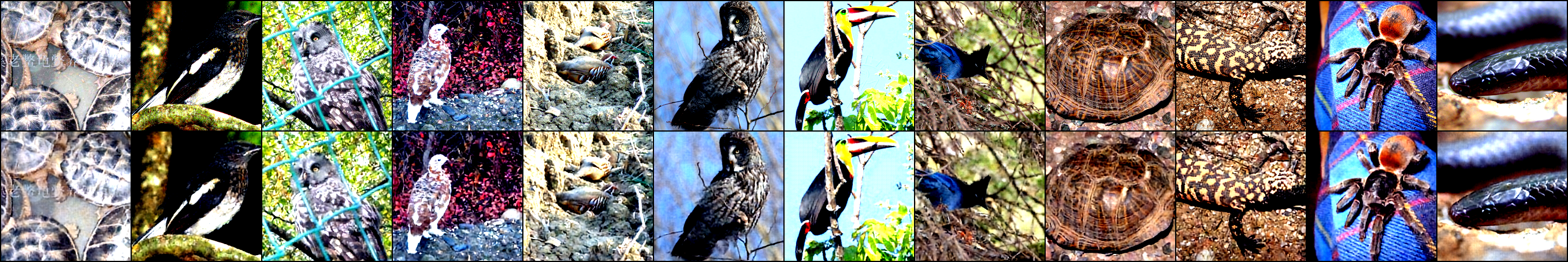}  \\
         (d) Reconstructed images from original ImageNet with resolution $224\times 224$\\
\end{tabular} 
\caption{Reconstructed images at different resolutions using a VQ-VAE pre-trained with 25,000 images ($32 \times 32$) from CIFAR-100.}
\label{fig:recon_img}
\end{figure*}

Example reconstructed images from CIFAR-100 are provided in Fig.~\ref{fig:recon_img}-(a). And those from ImageNet with resolution 32, 64, 224 are in 
Fig.~\ref{fig:recon_img}-(b), Fig.~\ref{fig:recon_img}-(c)), Fig.~\ref{fig:recon_img}-(d)) respectively.
Looking at the qualitative results presented in Fig.~\ref{fig:recon_img}, we argue that our VQ-VAE is capable of reconstructing varying sizes of images adequately, despite being applied to a dataset (ImageNet) that is different from the training dataset (CIFAR-100). To answer why the VQ-VAE has such a good generalization ability, intuitively, a VQ-VAE is just a pixel reconstructor and the data (50 classes from CIFAR-100) used for pretraining contains diverse images that provides a considerably good pixel distribution for learning.

\section{Compression for Codes via Bit-Swap}
Because we use the codebook size of $512$, the discrete value of a code ranges from $[0, 511]$. Theoretically, a code can be saved using 9 bits ($2^{9} = 512$), but in practice, a system saves codes data in the least unit of byte, so it causes 2 bytes, i.e., 16 bits to save an uncompressed code. According to~\textit{entropy coding} scheme, the entropy of a certain data distribution $p(\mathbf{x})$, defined by $H[\mathbf{x}] \triangleq \mathbb{E}[-\log p(\mathbf{x})]$, is the lower bound on how many average bits (called `message length') a lossless compression method can achieve to encode data points coming from this distribution. Table~\ref{tab:entropy} shows the entropy of top-level and bottom-level codes obtained by compressing a certain dataset via the VQ-VAE (the one used in our main paper). For example, the entropy of top codes and bottom codes of samples of $100$ classes from CIFAR-100 are $8.6032$ bits and $8.6404$ bits respectively. And Bit-Swap models trained by $3000$ epochs for each of them could achieve $8.8376$ or $8.8450$ average bits cost for compressing them respectively. 
We originally expected the more codes data (from $50$ classes to $100$ classes) can reveal a better distribution with less entropy in terms of a certain dataset. In contrast to our expectation, the entropy of codes distribution changed a little (less than 0.1) during the incremental setting on CIFAR-100, i.e., more new codes did not decrease the average overall entropy. Moreover, if we mixed the top and bottom codes, the entropy increased. We also ran preliminary results on Subset-ImageNet that contained random 100 classes ($128,856$ training samples) from ImageNet-1K with an image resolution of $224\times 224$. We obtained codes of ImageNet using the VQ-VAE pretrained on CIFAR-100 as discussed in the main paper. For all top-level codes of Sub-ImageNet, after trained for $2100$ epochs, a Bit-Swap cost $7.86$ average bits to encode one top-level code. As for bot-level codes, another Bit-Swap trained for $1100$ epochs cost $7.83$ average bits to encode one bottom-level code. Our preliminary results showed that more iterations (epochs) could further improve the compression performance.

\begin{table}[t]
    \centering 
    \begin{tabularx}{.45\textwidth}{l c c c c} 
    \hline
    Dataset  & top  & bottom  & top \& bottom \\ [0.5ex] 
    \hline

    CIFAR-100   & 8.6032 & 8.6404 & 8.7482 &\\
    
    SubImg 224x224 & 6.7936 & 7.1016 & 7.4683  \\

    \hline 
    \\
    \end{tabularx}
    \caption{Entropy of codes distributions for different dataset. `SubImg' refers to a subset (contains 100 classes) of ImageNet}
    \label{tab:entropy} 
\end{table}

\section{Implementation details of compared methods}
We compare our methods with the following baseline methods and state-of-the-art methods with their original implementation settings on CIFAR-100:  
  \begin{itemize}
    \item {\bf \textit{Upper Bound (UB)}}  saves all (50,000) exemplars and trains from scratch a Resnet-18 for each new training phase. For training UB, we use the same hyperparameters as our DRR.  
  \item {\bf \textit{GFR}}~\cite{CL_GFR} requires no exemplar to be stored. It has a two-stage training: The first stage jointly trains a feature extractor and a classifier for $200$ epochs, where they use a Resnet-18 and treat the former layers as the feature extractor. They save a copy of the old model after a training phase and initialize the new model with the weights of the old ones. The second stage trains a GAN used a frozen feature extractor and the classifier for $500$ epochs.
    \item  {\bf \textit{iCaRL}}~\cite{CL_iCaRL} saves $20$ exemplar per class and $2,000$ in total. It uses a rank function to select samples to be reserved, and a Resnet-32 is trained or fine-tuned for $160$ epochs.
     \item {\bf \textit{LUCIR}}~\cite{CL_LUCIR} saves $20$ exemplar per class using class-rebalance strategies. It also uses the rank function introduced in iCaRL. We use the setup of LUCIR as presented in~\cite{CL_Mnemonics} where a Resnet-32 is used and is fine-tuned for $160$ epochs in new training phases.
     \item  {\bf \textit{Mnemonics}}~\cite{CL_Mnemonics} is built upon LUCIR while it saves $20$ trainable exemplars per class instead. Two-level training includes model-level training for a Resnet-32 and exemplar-level training for exemplars. The classifier and learnable exemplars are fine-tuned in the training phase.
    \item  {\bf \textit{LWF}}~\cite{LWF} is one of the most representative non-replay-based methods and no exemplar is saved. It incorporates knowledge distillation technique~\cite{KNOWLEDGE_DISTILLATION} as an extra regularisation term to consolidate previous knowledge when learning new knowledge. We also use the setup of LWF as presented in~\cite{CL_Mnemonics} where a Resnet-32 is used and is fine-tuned for $160$ epochs in new training phases.

\end{itemize}

\section{The Choice of Resnet for CIFAR-100}
\label{subsec:resnet}
Residual Network (Resnet) was proposed by He~\etal~\cite{Resnet} in 2015. Resnet-18 was also proposed in~\cite{Resnet}, which had 18 layers with a hyperparameter `in-planes' set to 64. The `in-planes' was used to control the number of filters in layers. For example, with `in-planes = 64', the layers in $i_{th}$ ResBlock had $64 \times i$ filters. There were many variants of Resnet with a different number of layers and different `in-planes'. In recent incremental learning works researchers prefer a Resnet-32~\cite{CL_Mnemonics,CL_iCaRL,CL_CURL} with `in-planes = 16' that had more layers but lower capacity than original Resnet-18 with `in-planes = 64'. In our preliminary experiments, we found that Resnet-32 with `in-planes = 16' resulted in underfitting of our models in some cases, e.g., if a strong Data Augmentation was applied. Our further experiments showed that if we increased the number of filters in Resnet-32, i.e., setting the hyper-parameter `inplanes' from $16$ to $64$, our method performed even slightly better as compared to that used Resnet-18 with `inplanes=64'. 
We conjectured that Resnet-32 with `inplanes =16' was adequate only for replaying  $20$ (low resolutions) samples per class. However, for replaying more samples like ours and~\cite{CL_discreteVAE} or generated sample models (like GFR~\cite{CL_GFR}), Resnet-18 or Resnet-32 with `inplanes=64' was more suitable for CIFAR-100. Note that we only used simple data augmentation like `horizontal flip' and `random crop' for the experiments in the main paper, but our previous studies showed that DRR could benefit from a strong data augmentation strategy ~\cite{RL_ADA} as well as test-time augmentation strategy~\cite{TTA_1,TTA_2}. We will further test IB-DRR with the above data augmentation strategies in the future.

\end{document}